\documentclass[conference]{IEEEtran}
\IEEEoverridecommandlockouts
\usepackage{cite}
\usepackage{amsmath,amssymb,amsfonts}
\usepackage{algorithmic}
\usepackage{graphicx}
\usepackage{textcomp}
\usepackage{xcolor}
\usepackage{soul}
\usepackage{hyperref}
\def\BibTeX{{\rm B\kern-.05em{\sc i\kern-.025em b}\kern-.08em
    T\kern-.1667em\lower.7ex\hbox{E}\kern-.125emX}}
\begin{document}

\title{Universal Rules for Fooling Deep Neural Networks based Text Classification \\
{\footnotesize \textsuperscript{}}
\thanks{D. Li carried out the experiments and wrote the manuscript with support from D. V. Vargas. K. Sakurai helped supervise the project. D. V. Vargas conceived the original idea and supervised the project. 
Code: \href{https://github.com/ldxhdsz/research.git}{https://github.com/ldxhdsz/research.git} }
}
\author{\IEEEauthorblockN{Di Li}
\IEEEauthorblockA{\textit{Kyushu University} \\
Fukuoka, Japan \\
li.di.333@s.kyushu-u.ac.jp}
\and
\IEEEauthorblockN{Danilo Vasconcellos Vargas}
\IEEEauthorblockA{\textit{Kyushu University} \\
Fukuoka, Japan \\
vargas@inf.kyushu-u.ac.jp}
\and
\IEEEauthorblockN{Sakurai Kouichi}
\IEEEauthorblockA{\textit{Kyushu University} \\
Fukuoka, Japan \\
sakurai@inf.kyushu-u.ac.jp}
}

\maketitle

\begin{abstract}
Recently, deep learning based natural language processing techniques are being extensively used to deal with spam mail, censorship evaluation in social networks, among others. 
However, there is only a couple of works evaluating the vulnerabilities of such deep neural networks. 
Here, we go beyond attacks to investigate, for the first time, universal rules, i.e., rules that are sample agnostic and therefore could turn any text sample in an adversarial one. 
In fact, the universal rules do not use any information from the method itself (no information from the method, gradient information or training dataset information is used), making them black-box universal attacks. 
In other words, the universal rules are sample and method agnostic.
By proposing a coevolutionary optimization algorithm we show that it is possible to create universal rules that can automatically craft imperceptible adversarial samples (only less than five perturbations which are close to misspelling are inserted in the text sample).
A comparison with a random search algorithm further justifies the strength of the method.
Thus, universal rules for fooling networks are here shown to exist.
Hopefully, the results from this work will impact the development of yet more sample and model agnostic attacks as well as their defenses.
%In other words, we show here that it is also possible to craft universal rules that can fool deep neural networks with few perturbations for text.
%For that we propose an optimization algorithm to develop such rules. Our preliminary results show that more than 2000 modified texts are misclassified from 10000 modified samples. 
%Thus, the proposed attack shows 38\% of the samples can be fooled with such general rule, showing that the principle hold and further improvement on the rule and evolutionary algorithm may reveal further security issues.
\end{abstract}


\section{Introduction}
In recent years the fast development of deep learning (DL) brought great changes to techniques in many fields. 
Deep Neural Networks (DNNs) have been extensively applied in many fields, such as image recognition [7], computer vision [12], and natural language analysis [6]. 
In fact, DNNs achieve even human-competitive performance in many fields. 
Meanwhile the security problem of DNNs has become a critical topic.
\begin{figure}[htbp]
\centering
\includegraphics[width=9cm, height=13cm]{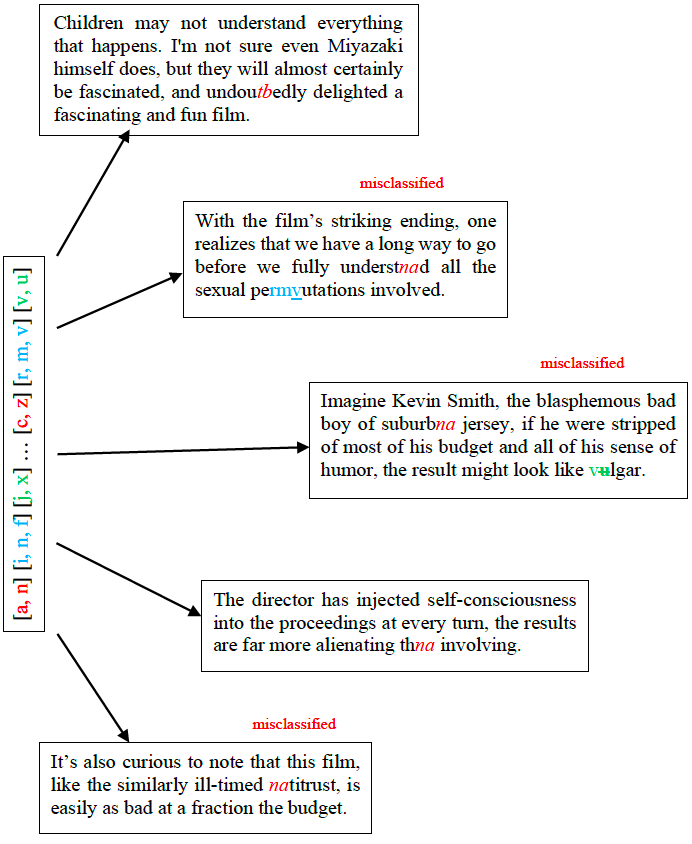}
\centerline{}
\caption{\textit{Example of a crafted universal rule (sequence of prototype-matching based perturbation procedures) for fooling text classification. 
Using the technique proposed in this paper it is possible to craft a universal rule which automatically create adversarial samples, i.e., once the universal rule is crafted no search is needed anymore.
In fact, the universal rule will only do a few perturbations which is imperceptible to typos.
In the figure, adversarial samples are generated by one universal adversarial rule (changes to the original sample are shown with a different color. Swapping is shown in \textcolor{red}{red}, deletion shown in \textcolor{green}{green}, insertion shown in \textcolor{blue}{blue}.).}}
\label{fig}
\end{figure}

Several recent studies [10, 11] demonstrate that some artificial perturbations can easily make DNN-based image or audio classifiers misclassify. 
Szegedy C. et al [34] first revealed the sensitivity to artificial perturbations. 
Specifically, the state-of-the-art GoogLeNet would misclassify the adversarial images generated by "fast gradient sign" algorithm [10], while a human observer can still classify correctly almost without noticing the artificial perturbations. 
These studies revealed the fact that the adversaries could potentially fool the state-of-the-art DNNs by crafting perturbations. 
Additionally, some researchers even investigated adversarial images created under extremely general or limited scenarios, such as universal adversarial perturbations [18] and one-pixel attack [28] for DNNs.

In the domain of text processing, DNN-based Natural Language Processing (NLP) could learn non-linear models, overcoming traditional NLP's linear model. 
Moreover, deep learning learns the language features itself without extracting, achieving higher precision. 
However, a recent study has revealed that artificial perturbations could also make DNN-based text classifiers misclassify. 
Unfortunately, DNNs for natural language processing have not got the attention they deserve and until recently previous attacks did not propose an effective algorithm for generating adversarial texts. 
Bin Liang et al. demonstrated that the text classification done with DNNs can also be attacked similarly to image or audio classification DNNs [2]. 
They successfully crafted adversarial samples for DNN-based natural language text classifiers.

Here, we propose a technique that aim to overcome some of the limitations of previous ones.
First, in [2] for locating Hot Training Phrases (HTPs) or Hot Sample Phrases (HSPs), the training dataset, features of data, dimensions of the model, classification items and some other information are necessary when crafting the adversaries. 
However, in practice, the conditions and data necessary are mostly unavailable. 
Second, there is a strong dependence on gradients which are not always available, i.e., in [2] if the cost gradients could not be computed, it is hard if not impossible to get the HSPs. 

In this paper, we propose a technique that can craft adversarial samples for a general black-box scenario.
In fact, our proposed method creates a universal rule that can create adversarial samples automatically, i.e., no search is necessary (Fig. 1 shows an overview of adversarial text samples crafted by a universal rule). 
Since the pertubation only change a few letters of a phrase, this attack is almost imperceptible.
The novelty of this work lies in proposing:
\begin{itemize}
\item {\bf Universal Rule} - The current approach is the first to create an automatic rule pattern that can process samples rapidly and output adversarial samples. 
This goes beyond universal perturbations to create yet another layer of abstraction which allows pertubations to change depending on the sample.
\end{itemize}

%and automatically generated attack.
%aft adversarial text samples under a general scenario, perturbing a deep text classifier by only one rule with high misclassify probability. 
Comparing to previous works our proposal has the following main advantages. 
\begin{itemize}
\item {\bf Automatic (Universal Perturbation)} - No need for searching for adversarial samples. The creation of an adversarial sample is done by a rule that changes some letters according to a learned pattern.
\item {\bf Black-box Attack} - There is no need to get any information of the target model and training dataset in advance; our approach could straightly act on an artificial test dataset. In fact, we use a metaheuristic to search instead of heuristically searching for important tokens \cite{b9}. 
\item {\bf Non-Gradient Method} - The proposed method does not compute or need to compute cost gradients. To create a universal rule, a novel black-box search algorithm is employed. 
\item {\bf Imperceptible} - Every sample phrase can only be perturbated five times or less. Thus the resulting sample is imperceptible to human observers.           
\end{itemize}

\section{Related Work}   

With the development of DL, DNNs have been widely applied in many fields an therefore DNNs' security problem came to be of utmost importance. 
There have been many works investigating the security of DNNs as well as identifying its vulnerabilities by proposing several attack methods [1], including black-box attacks [26] and the white-box attack [15]. 
Various methods and algorithms are proposed to generate adversarial samples, including gradient-based (e.g. "fast gradient sign" algorithm proposed by I.J.Goodfellow et al.) [10, 11, 19], greedy approaches (e.g. greedy perturbation searching method proposed by S.M. Moosavi-Dezfooli et al.) [21, 17] , and evolution-based (e.g. one-pixel attack proposed by Su Jiawei et al.) [19, 28].                                                                                                                                                                                                                                               

Some researchers consider that the state-of-the-art deep neural networks are highly vulnerable to gradient-based methods which is easy to use as well. 
For instance, in recent years Moosavi D. et al. proposed systematic algorithm [18] for computing universal perturbations which caused natural images to be misclassified with high probability. 
In addition, Su Jiawei et al. proposed one-pixel attack by using a differential evolution to search under an extreme limited scenario.
Therefore, there are many types of optimization methods which result in high misclassified rates for DNN-based image classifiers. 

Unfortunately, there are no studies paying attention to methods or algorithms for generating universal perturbations against DNN-based text classification. 
Text as a discrete data is also sensitive to perturbation, however, the geometric correlations among the high-dimensional decision boundary of classifiers couldn't be found in text data, so the existing algorithms for generating adversarial images cannot be directly applicable for text. 
%We need to investigate new attacks and corresponding defenses.  
Recently, Bin L. et al. proposed the first attack for deep text classification[2]. 
%Thebased gradient-based perturbation methods 
Additionally, Bin Liang et al. demonstrated that since text is a kind of discrete data, when directly adopting existing image or audio perturbation algorithms the resulting text sample may lose its original meaning or even become meaningless for human observers [2]. 
Thus, in order to craft imperceptible adversarial text samples without losing their original meaning, they presented three perturbation strategies: insertion, modification, and removal.
To craft them, they used the cost gradients for original text and training samples to generate adversarial samples.

In order to maintain the meaning of a text sample, they perturbated the sample by directly modifying its words, inserting new items (words or sentences) or removing some original ones from it. 
First, for all training samples the cost gradients of every dimension in all character vectors are calculated.
They termed phrases with significant cost gradients to the current classification as HSPs.
Additionally, the most frequent phrases in all training samples of the target classification are termed as HTPs. 
For insertion strategy, HTPs form target classifications are inserted into the text samples nearby phrases with significant contribution to the original class which result in the increase in confidence of target class and decline in original classification confidence. 
In modification and removal perturbations, HSPs for original classification are modified or removed which could generate the drop off in original confidence. 
Fig. 2 presents an example of the three proposed perturbation strategies.        

\begin{figure}[htbp]
\centering
\includegraphics[width=8cm, height=3cm]{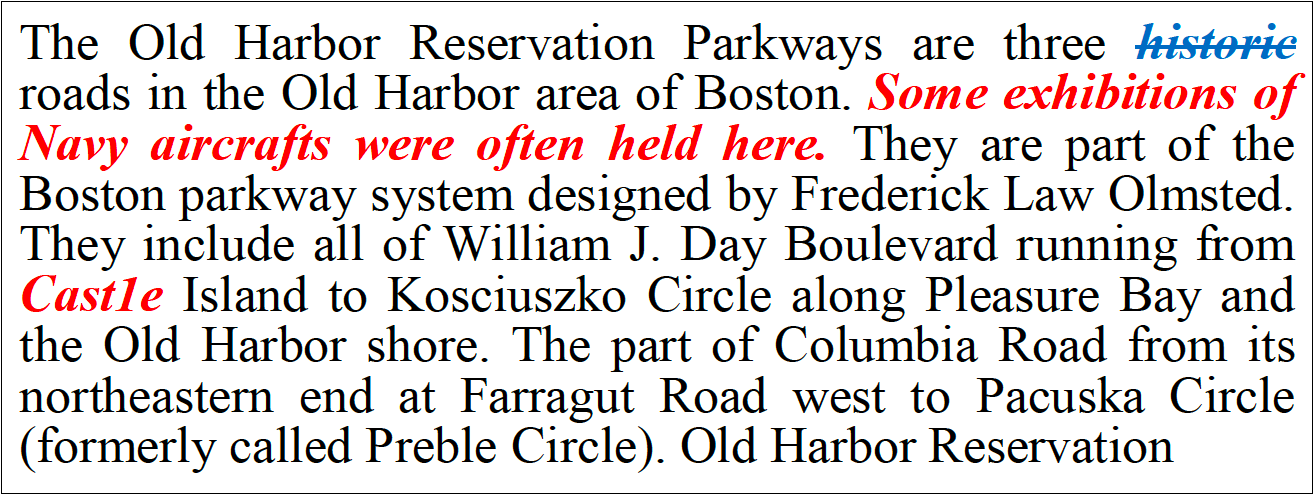}
\centerline{}
\caption{\textit{An adversarial text sample generated with Bin Liang et al. [2] proposed perturbations. 
Inserting a gorged fact: \textcolor{red}{Some exhibitions of Navy aircrafts were often held here.}, removing an HTP: \textcolor{blue}{historic}, and modifying an HSP: \textcolor{red}{Castle}. The output classification is successfully changed from Building to Means of Transportation.}}
\label{fig}
\end{figure}  

\section{Target Model and Settings}

Convolutional Neural Networks (CNNs) which is typically used in computer vision can also be applied to problems in Natural Language Processing and perform quite well. 
Location invariance and local compositionality made intuitive sense for images but not so much for NLP.
Considering all this, it seems like CNNs wouldn't be a good fit for NLP tasks. 
However, CNNs are fast and efficient in terms of NLP tasks as well because they can extract relationships from words and sequences.

A recent study investigates the use of CNNs to learn directly from characters without the need for any pre-trained embeddings [35]. 
However, results show that learning directly from character-level input works very well on large datasets (millions of samples) but underperforms in simpler models on smaller datasets (hundreds of thousands of samples). 
Therefore, we apply a common CNN for text classification [3]. 
For model training and evaluation, we use the same as in [3] which employs the Movie Review data from Rotten Tomatoes.
This dataset contains $10,662$ movie review sentences, half positive and half negative. 
Positive and negative sentences are loaded from the raw data files and cleaned for feeding input texts into the network.
Additionally, instead of using the pre-trained word2vec vectors for word embeddings, this model directly learn embeddings from scratch. 
In other words, the first layer words are embedded into low-dimensional vectors and afterwards the next layer performs convolutions over the embedded word vectors using multiple filter sizes.
This is followed by a max-pooling over time and a last layer of fully-connected neurons with dropout regularization which outputs in a softmax layer (Figure 3 shows the models for the two types of DNN models used).    
%\hl{Only one model is explained..}

\begin{figure*}[htbp]
\centering
\includegraphics[width=14cm, height=8cm]{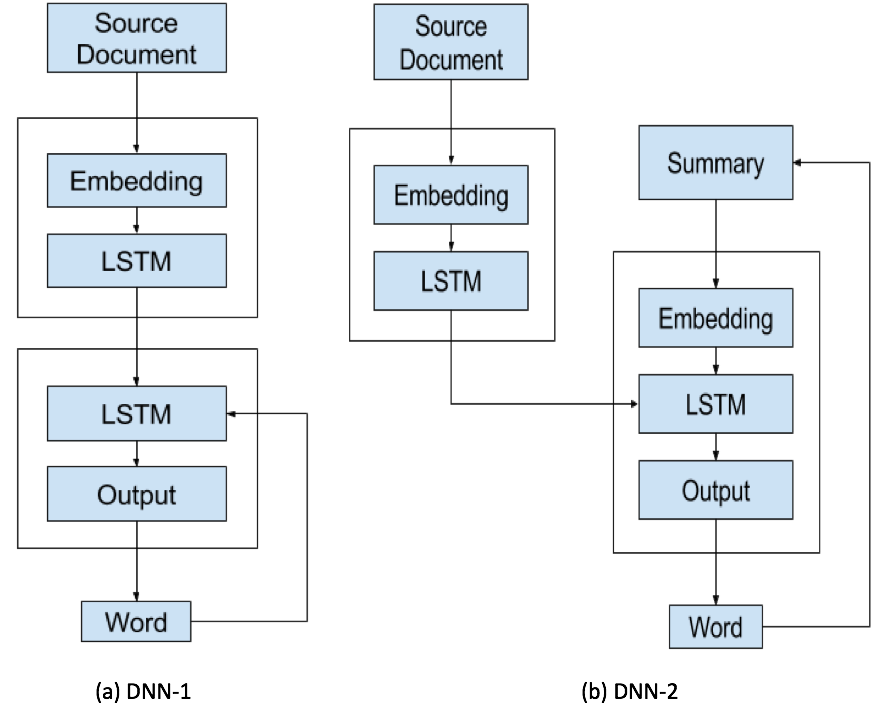}
\centerline{}
\caption{\textit{Applied Convolutional Neural Network models for text classification (DNN-1 and DNN-2) [3]}}
\label{fig}
\end{figure*}

%In our research, we apply two deep learning based natural language processing base on the above model for experiments.

\section{Universal Rule}
\label{ur_sec}

The proposed method is both a random search and a coevolutionary optimization algorithm for generating universal rules that can create adversarial samples for DNN-based natural language text classifier [3]. 
Before introducing the optimization algorithm, the three perturbation procedures of which the universal rule is consisted of are explained in detail.
The universal rule is made of a sequence of perturbation procedures which can be either swapping, deletion or insertion procedures (the whole procedure is illustrated and explained in Fig. 7).
Each of these procedures are described in detail below.
The universal rule itself is made of many perturbation procedures in sequence.
However, that does not mean a text will be perturbed many times because there is a limit to the number of pertubations set to five.
Note that although this limit is set, in a given sample most of the time there will be three or less perturbations.
This happens because it is hard to find matching letters for each rule.
In fact, we found that three perturbation methods performed not well in practice when used separately.
Using only one such perturbation for attacking could get $3\%$ fooling rate at most because many of the rules cannot find a pattern that matches them.
Therefore they fail to modify the sample not mention make it misclassify. 
Thus, we use $10$ or more different perturbations to make up a universal rule.

\subsection{Swapping Perturbation Procedure}

%For a given text \textit{t}, the goal of the method is to make \textit{t} to be misclassified by swapping two letters of a word in \textit{t}. 
%Because we perform our attacks without using the training dataset and target model, the letters for a swapping perturbation are randomly created. 
%In fact, the letters might not exist in the samples.
%In this case, the perturbation would be ignored. 

A swapping perturbation procedure is defined by two letters.
These are the letters that will be searched for among words in the original text sample.
Once it is located in a word, the two letters in the word will be swapped and the word becomes a slightly misspelled word.
The results of our test show that even such imperceptible change could make the DNN text classifier misclassify.
\begin{figure}[htbp]
\centering
\includegraphics[width=7.5cm, height=2.1cm]{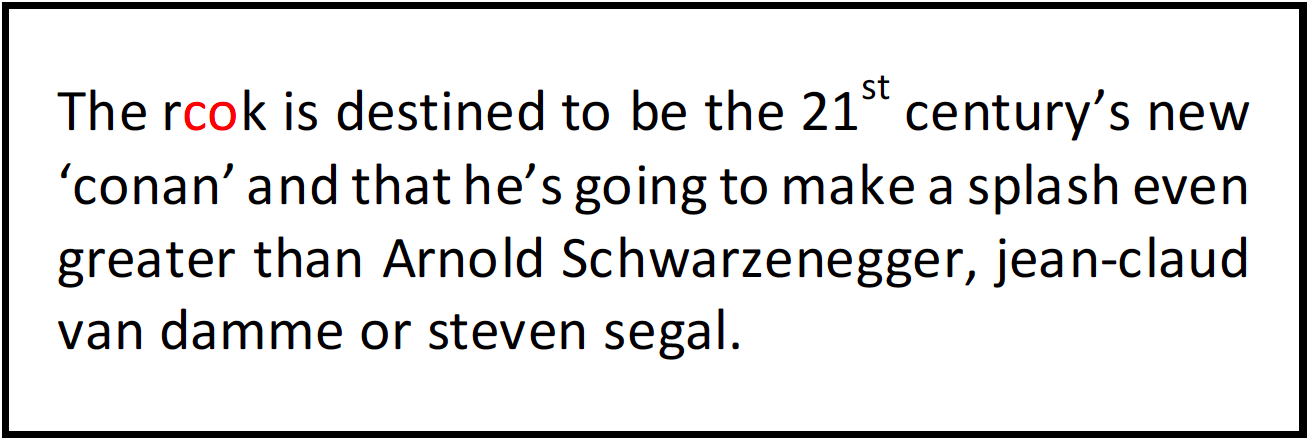}
\centerline{}
\caption{\textit{An adversarial text sample generated by swapping the two letters (two elements of perturbation, shown in  \textcolor{red}{red}).}}
\label{fig}
\end{figure}

For example, Fig. 4 shows a text sample classified as \textit{positive review} class. 
Swapping letters of the word in the sample just one time, the perturbed sample is classified as \textit{negative review} class. 
However, for human observers, we still can recognize the text as a positive one and even know the misspelled word is \textit{rock}.

\subsection{Deletion Perturbation Procedure}

The deletion perturbation procedure deletes a letter from matched word in a sample.
%W still randomly generate perturbations without caring the model's sensitivity. 
This procedure is defined by two letters.
Given these two letters, the procedure searches for them over the text and once it founds a match the second letter is deleted from the word.
Fig. 5 shows an example, the deletion perturbation is \textit{[o, o]}, after deleting the second \textit{o} the text's classification changed from \textit{negative review} to \textit{positive review}.
\begin{figure}[htbp]
\centering
\includegraphics[width=4cm]{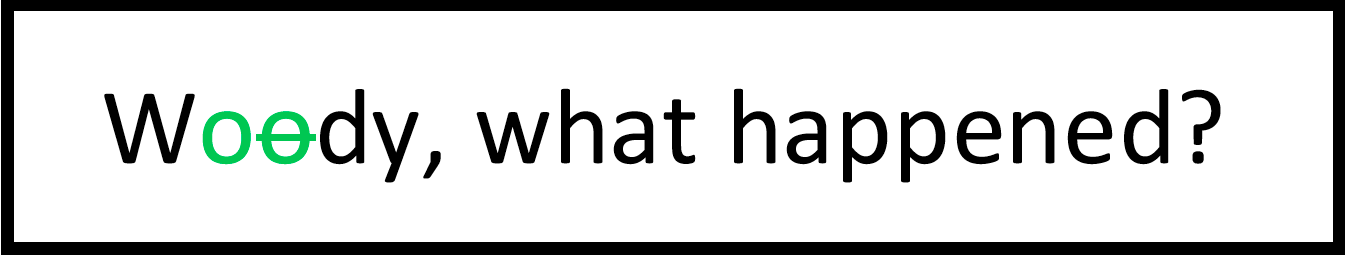}
\centerline{}
\caption{\textit{An adversarial text sample generated by deletion method, the second perturbation element  \textcolor{green}{o} is deleted after being located (shown in  \textcolor{green}{green}).}}
\label{fig}
\end{figure}

%Using only one deletion perturbation is hard to perturb all the text samples from test dataset or get a decent fooling rate. 
%Therefore, we generate five perturbations for one set and use the set to perturb $10,000$ samples gaining average 1.5\% fooling rate. 

\subsection{Insertion Perturbation Procedure}

The insertion perturbation procedure perturb the classification probability by inserting a letter into a word.
The misspelled word might lead to the decrease of the original class confidence or the increase of the miss-class confidence. 
This procedure consists of three letters which works by searching for the first two letters in a word over the text.
When a match is found the third letter is inserted after them. 
In Fig. 6 an example is presented in which two letters \textit{i} and \textit{l} are located in the word film, then it is perturbed into a misspelled word \textit{filam} and results in misclassification.
\begin{figure}[htbp]
\centering
\includegraphics[width=8cm, height=1.5cm]{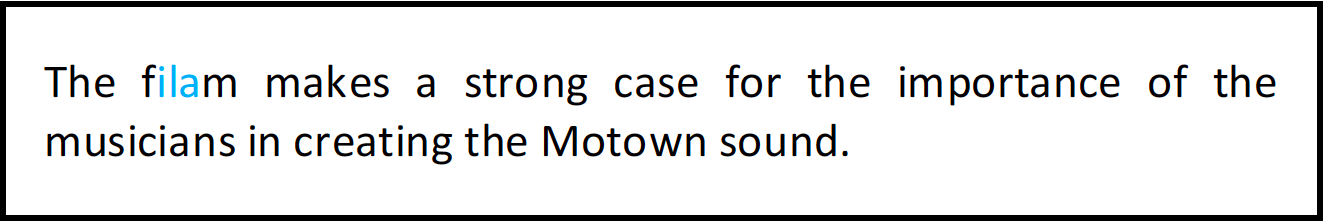}
\centerline{}
\caption{\textit{An adversarial text sample generated by inserting the third element  \textcolor{blue}{a} of the perturbation [\textcolor{blue}{i}, \textcolor{blue}{l}, \textcolor{blue}{a}], when it is located in the original word \textit{film}.}}
\label{fig}
\end{figure}

\begin{figure}[htbp]
\centering
\includegraphics[width=8cm, height=7cm]{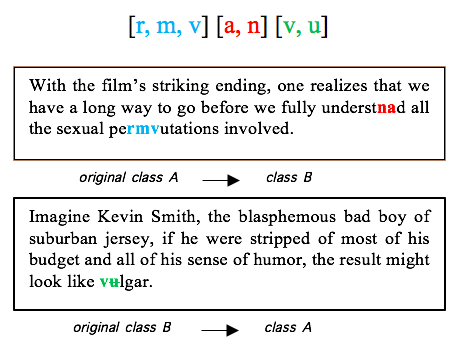}
\centerline{}
\caption{\textit{Example of a universal rule containing one of each perturbation procedure, i.e., insertion, swapping and deletion. 
For every text sample, the universal rule applies the perturbation procedure which appears first.
In this case an insertion procedure.
If it could not find a match it then continues with the next perturbation procedure in sequence.
Perturbations that can match and modify the sample are counted and if the maximum of perturbation is reached the modification (attack) ends.
If the universal rule reaches the end of the sequence then the modification (attack) also ends. 
}}
\label{fig}
\end{figure}  

\section{Universal Rule Evaluation}

To develop universal rules it is necessary to create some ways to evaluate them.
Here, we propose two types of fitness function.
One based on accuracy and the other based on utility and therefore the names: accuracy fitness and utility fitness.

Regarding the accuracy fitness, it is defined as the success rate of the attack from all universal rules created with the individual.
Notice that this measure also depends on the sample and the other individuals that take part on the universal rule set.
Thus, given the attack success of a given sample $AttackSuccess_i$ ($1$ when the class changed and $0$ otherwise), the accuracy fitness $F_a$ can be obtained by:
\begin{equation}
F_a = \frac{1}{n}\sum_i AttackSuccess_i,
\end{equation}
where $n$ is the number of times a given individual is evaluated. 
Notice that $n$ may vary from individual to individual since they are randomly picked each time a universal rule is created.

Regarding the utility fitness, universal rules should perturbate the sample a given number of times.
Although the maximum number of times that it can perturbate is set, there is no guarantee that it will perturbate this number of times.
Moreover, many individuals which code perturbation procedures fail to perturbate in most of the samples.
To avoid inactive individuals, a utility fitness is defined in which the value is the number of times an individual perturbate divided by the number of times it was chosen to participate in a universal rule. 
Therefore, considering $F_u$ the utility fitness and $Perturbate_i$ a variable that is either $1$ when sample $i$ is perturbated or $0$ otherwise, the following equation defines it explicitly:
\begin{equation}
F_u = \frac{1}{n}\sum_i Perturbate_i,
\end{equation}

\section{Random Search for Universal Rule Optimization}

In this paper, we propose two methods to develop universal rules.
The first one is a simple method called random search for universal rule optimization (RS).
The method consists of a variation of a random search procedure in which the best universal rule found is stored and returned as the output.
Specifically, the individuals of the population are made of perturbation procedures of the types described in Section~\ref{ur_sec}.
In each generation, first a new set of individuals is generated.
Afterwards, $100$ universal rules are created by combining the individuals into sequences.
Lastly, new individuals are created by mixing the individuals of the population using a differential operator like rule.
Since, the initial individuals are randomly spreaded inside the hypercube of possible perturbation procedures, the differential evolution operator will create random walks in this space. 

%\begin{table}
%\centering
%\begin{tabular*}{\hsize}{lrrrrr} \hline
%\textbf{Algorithm 1} DE for Universal Rules Optimization\\ \hline
%   \;\;1:  Random generation of individuals (parents and children)\\
%   , each encoding a perturbation procedure. \\
%   \;\;2: \textbf{Generation Loop}\\
%   \;\;3: \quad \quad \textbf{while} Universal Rules Generated $< 100$ \textbf{do}\\
%   \;\;4: \;\;\quad \quad\quad \quad Generate new Universal Rule from individuals\\ 
%   \;\;4: \;\;\quad \quad\quad \quad Evaluate generated Universal Rule \\ 
%   \;\;4: \;\;\quad \quad\quad \quad Update individual rank based on  \\ 
%   \;\;5: \quad \quad \textbf{end while}\\
%   \;\;6: \quad \quad Select .\\
%   \;\;7: \quad \quad Generating 100 new candidates (children set) by an evolutionary\\ 
%           \quad \quad \quad \quad parameter and father set. \\ 
%   \;\;8: \quad \quad Replacing father set with children set for next generation.\\
%   \;\;9: \textbf{END GENERATION}\\ \hline
%\end{tabular*}
%\end{table}

\section{Coevolutionary Algorithm for Universal Rule Optimization}

Here we propose the Coevolutionary Algorithm for Universal Rule Optimization (CAURO).
The aim of the method is to find universal rules efficiently based on the combination of useful and accurate small perturbations rules.
Since the number of perturbation procedures is not fixed as well as the order and permutation of these perturbation procedures are also as important as the perturbation procedures themselves, coevolution seems to be a good match.
The objective here is to focus more on the combination of good rules rather than on creating one.
Moreover, we hypothesize that the optimization landscape for searching for universal rules is not a well behaved space to search, since good universal rules might be far away from each other with many less good solutions around.

\begin{figure*}[htbp]
\centering
\includegraphics[width=1.8\columnwidth]{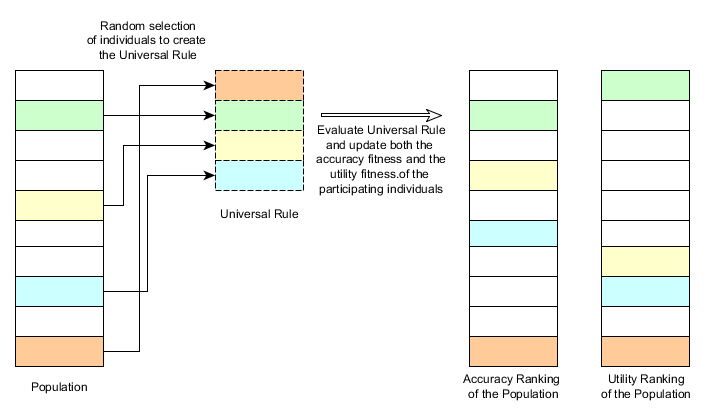}
\centerline{}
\caption{\textit{Evaluation of universal rules.  
For every rule, $k$ individuals encoding perturbation procedures are randomly selected the population.
Afterwards, the universal rule perturbs the dataset following the sequence of perturbation procedures it is made of.
Given the accuracy and utility fitness, the individuals evaluation is updated together with their ranking.
This process is executed many times until a certain number of universal rules are generated and consequently many individuals are already evaluated.
}}
\label{figeva}
\end{figure*}

\begin{figure*}[htbp]
\centering
\includegraphics[width=1.6\columnwidth]{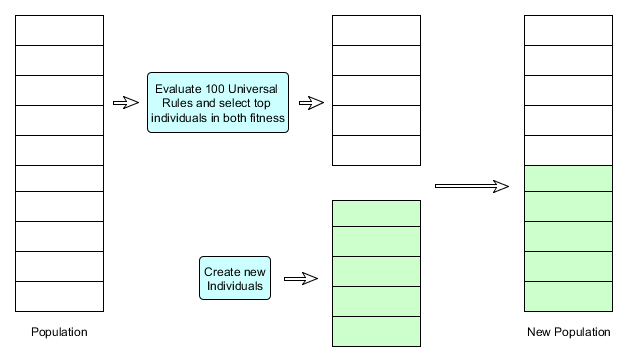}
\centerline{}
\caption{\textit{Overall illustration of the CAURO method proposed.
Since CAURO is a coevolutionary method the individuals do not encode the solution but rather building blocks of the solution.
Every generation universal rules are generated many times, individuals are evaluated.
Afterwards, the most fitted ones are selected to continue to the next generation while new individuals are randomly generated to fill up the population.
}}
\label{figover}
\end{figure*}

The algorithm consist of generating perturbation procedures to compose a population at first and then randomly picking individuals to compose universal rules.
Each time a universal rule is picked, it is evaluated in the dataset and have its constituting individuals update both its accuracy and utility fitness.
This evaluation process is illustrated in Figure~\ref{figeva}.
After a universal rule is created many times, the algorithm will rank individuals by accuracy and utility fitness.
This is followed up by a simple selection process in which $20\%$ of the individuals with lowest accuracy fitness and $20\%$ of the individuals with lowest utility fitness are replaced by new random generated individuals (Figure~\ref{figover}).
This composes the new population for the next generation.

\section{Evaluation and Results}

In this section, we evaluate the effectiveness of the proposed attack method by comparing the rate of misclassification of the best universal rule generated per generation.
The experiments are conducted on the movie review dataset from rotten tomatoes with two different DNN models. 
%We assign different values to two variables (i.e., rule size, maximum number of times a sample can be modified, perturbation methods) for both of the models to measure the performance of the proposed algorithm. 
Moreover, if most of the perturbation procedures in a universal rule are able to perturbate a text sample, they may modify a sample too many times.
This might cause the text sample to result in a meaningless sentence even for human observers.
In fact, since such an attack alters the content of the text sample it is considered a failed attack.
To avoid this problem, we decide that a sample can only be modified five times at most by a universal rule.
%Five misspelled words in a text is considered normal by human observers.

\begin{figure}[htbp]
\centering
\includegraphics[width=9cm, height=4.5cm]{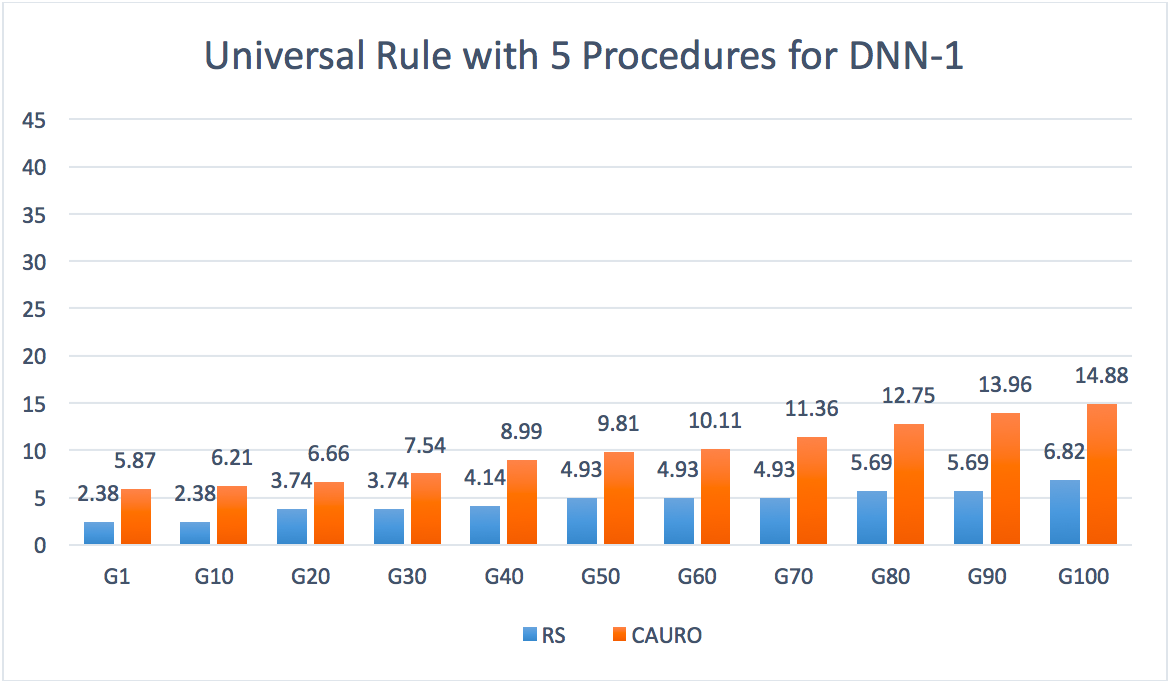}
\includegraphics[width=9cm, height=4.5cm]{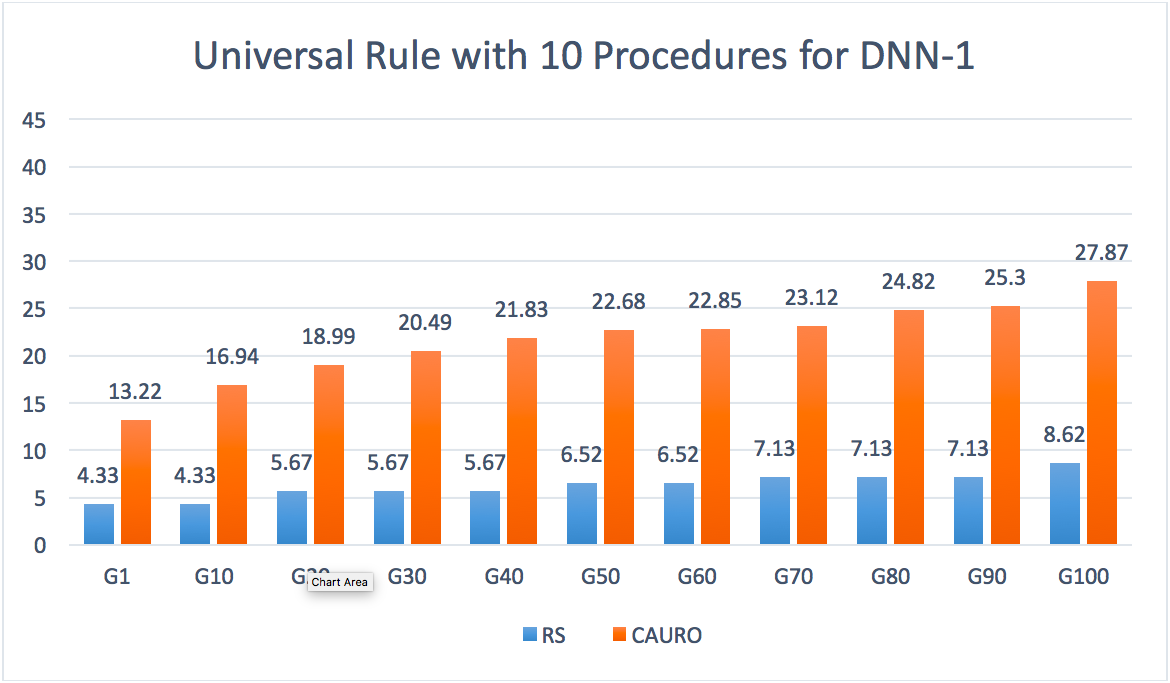}
\includegraphics[width=9cm, height=4.5cm]{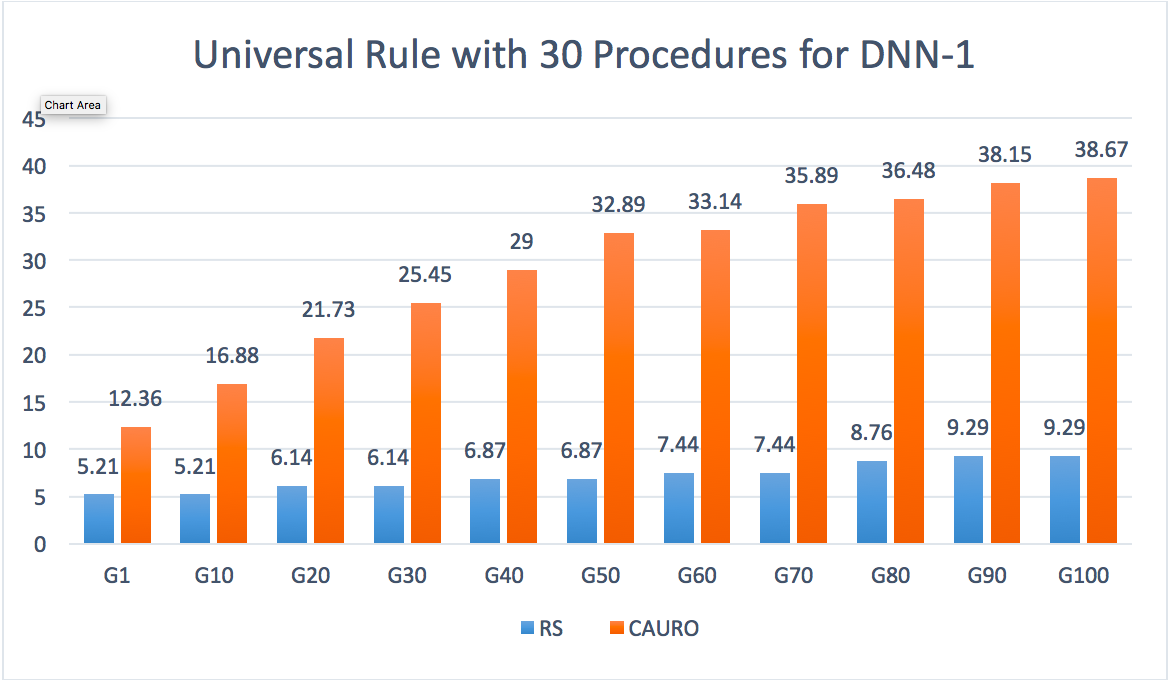}
\centerline{}
\caption{\textit{Misclassification rate in \% (ordinate) per generation (abscissa) for both the CAURO and RS while attacking DNN-1. 
Each graph shows the evolution of universal rules with a different sequence of perturbation procedures ($5$, $10$ and $30$).
However, the \textbf{maximum number of perturbation per sample is fixed to five.}
}
%The three charts show best success rates of deep text classification perturbed by 1 or 3 methods and every universal rule contains various perturbations. 
%Blue bar shows using differential evolution algorithm (DE) for optimizing universal rules, orange bar shows using the proposed CAURO for optimization. 
%Obviously, compare with DE, CAURO is much more effective for optimizing the presented universal rules.
}
\label{comp1}
\end{figure}

\begin{figure}[htbp]
\centering
\includegraphics[width=9cm, height=4.5cm]{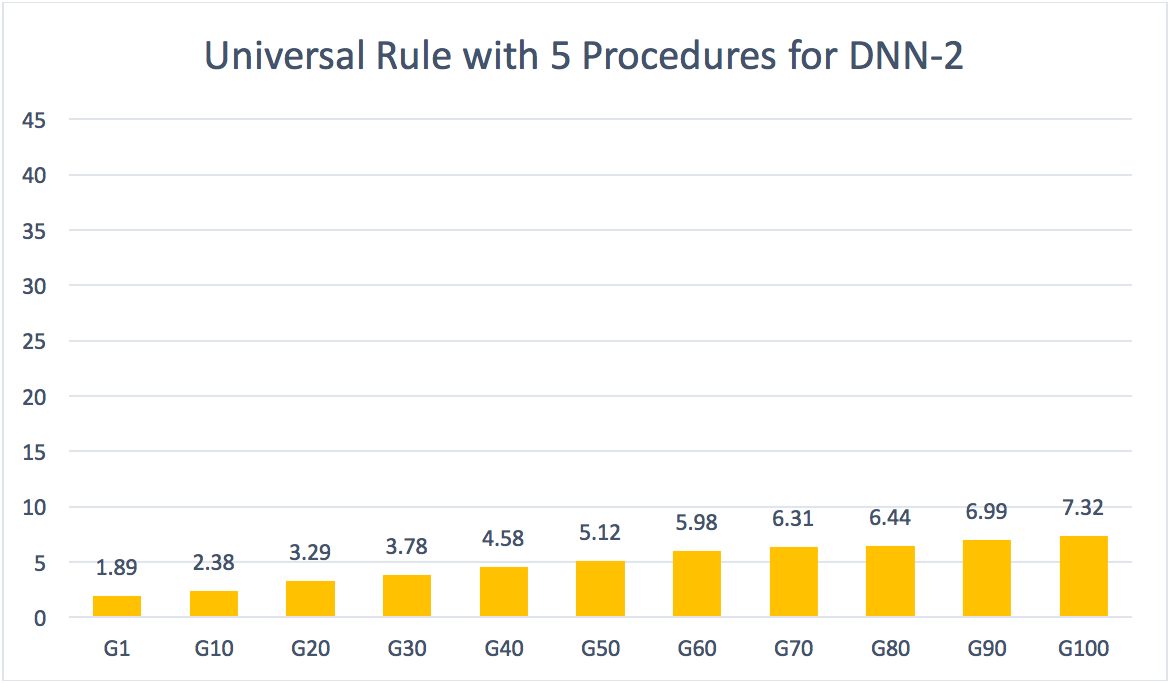}
\includegraphics[width=9cm, height=4.5cm]{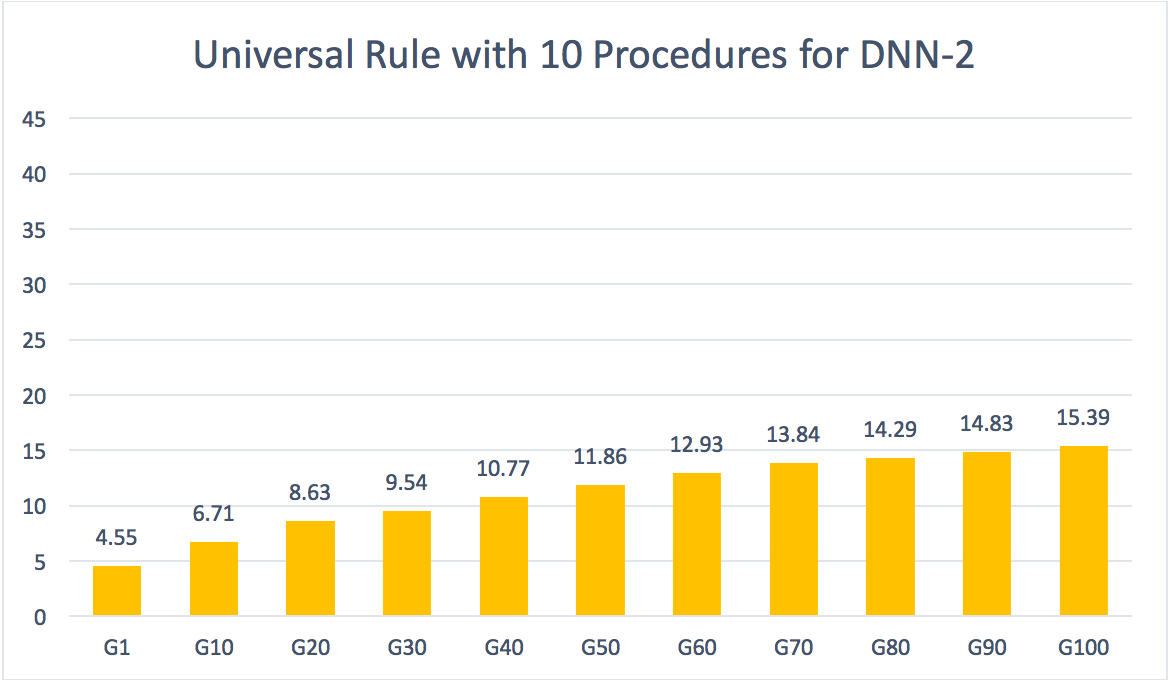}
\includegraphics[width=9cm, height=4.5cm]{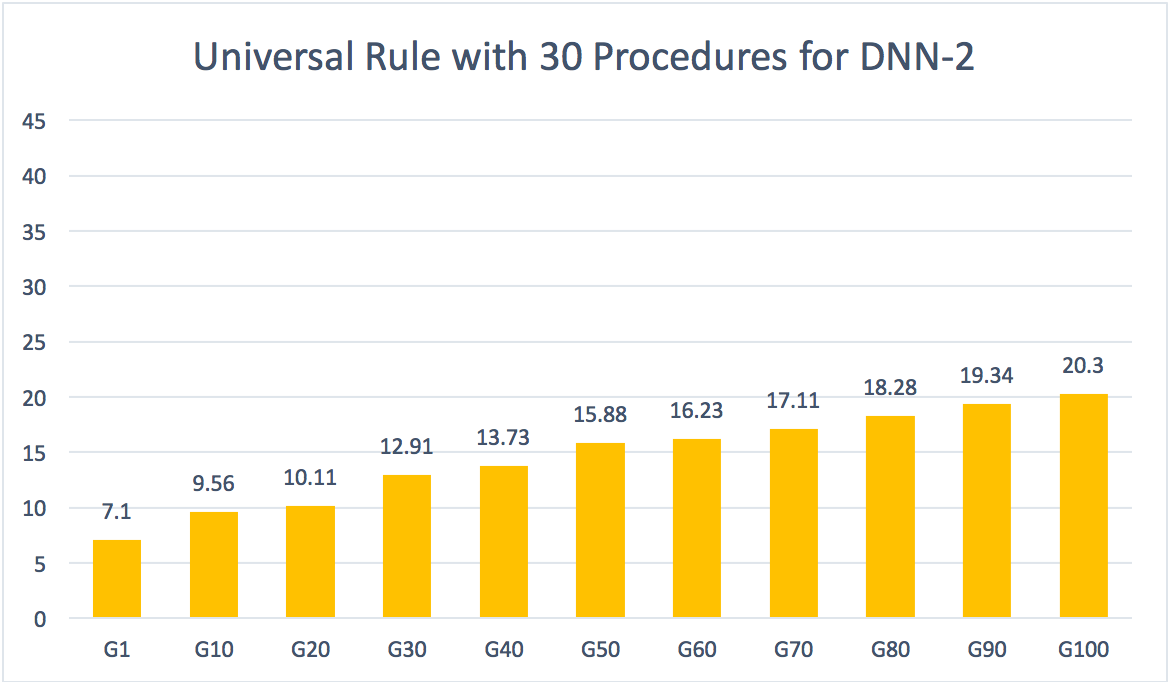}
\centerline{}
\caption{\textit{Misclassification rate in \% (ordinate) per generation (abscissa) for CAURO while attacking DNN-2.
Each graph shows the evolution of universal rules with a different sequence of perturbation procedures ($5$, $10$ and $30$).
However, the \textbf{maximum number of perturbation per sample is fixed to five}.
}}
\label{comp2}
\end{figure}

In Figure~\ref{comp1}, the misclassification rates of the best generated universal rules by RS and CAURO are shown.
After $100$ generations, the best misclassification rate of RS reached $9.29\%$ while CAURO achieved $38.67\%$.
In other words, CAURO creates universal rules which can fool DNNs with higher accuracy than RS.
Moreover, with the increase in the number of perturbation procedures, misclassification rates continues to increase. 
Notice that although the number of perturbation procedures in one universal rule increases, the maximum number of perturbation is fixed and kept at five per sample for all tests.
The performance of the proposed method CAURO is clearly superior to RS.
This shows that selection for better perturbation procedures as well as the utility and accuracy fitness are important to find better universal rules.
In Figure~\ref{comp2}, additional results are shown for the same dataset but with a different DNN (DNN-2 is used, for the description of the model please refer to Figure 3).
This time the performance decreases but it is still capable of achieving more than $20\%$ success rate after 100 generations.

Regarding the number of perturbations per rule, the reason for the low misclassification rates when universal rules are small ($5$ or $10$ perturbations in one rule) can only be justified by the fact that matching rules is a difficult process.
In fact, by looking at the data for the $30$ perturbations in one rule case, we verified that even with $30$ perturbations CAURO's crafted universal rules only modify $94\%$ of the test samples with on average $26\%$ of the text samples being modified five times.
Therefore, even for the case in which more perturbations are encoded in one universal rule, $74\%$ of the text samples are still perturbated less than five times.
%timesEven $94\%$ is a low number considering that there are $30$ perturbations procedures.
In other words, the experiments show that rules will often fail to match the text sample, resulting in a non perturbated one.
Moreover, with few perturbation procedures in one rule (e.g., $5$ or $10$ perturbations in one rule), matching rates possibly becomes more important than misclassifying rate.
This happens because every single perturbation procedure that matches and perturbs, increase the misclassification rate.
Consequently, individuals encoding perturbation procedures with rare but accurate perturbations should not survive in the such populations.

Thus, it is possible to create universal rules that can create adversarial samples without the need to search for them.
Actually, $38\%$ and $20\%$ success rate might not seem much at first glance.
However, this is not an adversarial attack success rate but rather a universal rule success rate.
This means that once an universal rule is found, no search is necessary for a given sample to become an adversarial sample because they are generated by just applying a simple universal rule made of a sequence of perturbation procedures.
We point out that the strong representation power of state-of-the-art neuroevolution methods using unified neural models [32] and the adaptiveness of self-organizing classifiers (which can adapt to changes in mazes similar to rats) [33] and MAP elites (which can adapt to malfunctions).

\section{Conclusion and Discussion}

Previous research has shown that DNN-based text classification is also vulnerable to the gradient-based adversarial samples. 
In this paper, we show the existence of universal rules (perturbations created from rules which are sample agnostic) that can fool state-of-the-art text classifiers. 
%In fact, we proposed two optimizing algorithms to generate universal rules.

In summary, this paper has the following main achievements:
\begin{itemize}
\item Universal Rules - We have shown that it is possible to generate universal rules that are sample agnostic, i.e., rules that can create adversarial samples without any search and independent of the sample given. 
\item CAURO - The proposition of a coevolutionary algorithm (CAURO) for generating universal rules efficiently. 
In fact, it is the first time that a coevolutionary algorithm is applied to adversarial machine learning. 
\end{itemize}

The results achieved here should impact new adversarial attacks as well as their defenses.
CAURO can be extended to other types of input such as images as well as can incorporate other types perturbation procedures, e.g. the repetition of words in the text or even more complex forms of perturbation.
Adversarial samples used here can be used to investigate the reason of the attacks and their respective defenses.
Moreover, we expect our work to also incentivate new methods that could themselves, without any kind of specifically designed defenses, overcome the current limitations.

\section{Acknowledgement}

This work was supported by JST, ACT-I Grant Number JP-50166, Japan.
%We would like to thank the reviewers for helping improve this paper with their valuable comments. 

\end{document}